\documentclass[11pt]{article}

\usepackage{acl}

\usepackage{times}
\usepackage{latexsym}

\usepackage[T1]{fontenc}

\usepackage[utf8]{inputenc}

\usepackage{microtype}

\usepackage{inconsolata}
\usepackage{microtype}
\usepackage{todonotes}
\usepackage{booktabs}
\usepackage{paralist}
\usepackage{colortbl}
\usepackage{amsmath}
\usepackage{blkarray, array}
\usepackage{framed}
\usepackage{subcaption}

\usepackage{xcolor}
\usepackage{hyperref}
 \definecolor{darkblue}{rgb}{0, 0, 0.5}
 \hypersetup{colorlinks=true, citecolor=darkblue, linkcolor=darkblue, urlcolor=darkblue}
\usepackage{multirow}
\usepackage{xstring}

\usepackage{color}
\usepackage{graphicx}
\usepackage{tabularx}
\usepackage{soul}
\usepackage{mathtools}
\usepackage[capitalise,noabbrev]{cleveref}
\usepackage{xcolor}
\usepackage{placeins}

\newcommand\citex[1]{\textit{\citeauthor{#1}}}

\DeclareMathOperator{\multi}{m}
\DeclareMathOperator{\bin}{b}

\DeclareMathOperator{\full}{full}
\DeclareMathOperator{\pbin}{b*}
\DeclareMathOperator{\pmulti}{m*}
\DeclareMathOperator{\rf}{RF}

\title{Computational Approaches for Integrating out Subjectivity in Cognate Synonym Selection}

\author{
  Luise Häuser \\
  Computational Molecular Evolution group, \\
  Heidelberg Institute for Theoretical\\
  Studies, Heidelberg, Germany, \\
  Institute for Theoretical Informatics, \\
  Karlsruhe Institute of Technology, Karlsruhe, Germany \\
  \texttt{luise.haeuser@h-its.org} \\
  \And
  Gerhard Jäger \\
  University of Tübingen \\
  \texttt{gerhard.jaeger@uni-tuebingen.de} \\
  \AND
  Alexandros Stamatakis\\
  Biodiversity Computing Group, \\
  Institute of Computer Science, Foundation for Research and Technology - Hellas\\
  Computational Molecular Evolution group, \\
  Heidelberg Institute for Theoretical Studies, Heidelberg, Germany
  Institute for Theoretical Informatics, \\
  Karlsruhe Institute of Technology, Karlsruhe, Germany \\
  \texttt{stamatak@ics.forth.gr}
}

\begin{document}
\maketitle
\vspace{5cm}
\begin{abstract}
Working with cognate data involves handling synonyms, that is, multiple words that describe the same concept in a language. In the early days of language phylogenetics it was recommended to select one synonym only. However, as we show here, binary character matrices, which are used as input for computational methods, do allow for representing the entire dataset including all synonyms. Here we address the question how one can and if one should include all synonyms or whether it is preferable to select synonyms a priori. To this end, we perform maximum likelihood tree inferences with the widely used RAxML-NG tool and show that it yields plausible trees when all synonyms are used as input. Furthermore, we show that a priori synonym selection can yield topologically substantially different trees and we therefore advise against doing so. To represent cognate data including all synonyms, we introduce two types of character matrices beyond the standard binary ones: probabilistic binary and probabilistic multi-valued character matrices. We further show that it is dataset-dependent for which character matrix type the inferred RAxML-NG tree is topologically closest to the gold standard. We also make available a Python interface for generating all of the above character matrix types for cognate data provided in CLDF format. 
\end{abstract}

\section{Introduction}
Lexical data are frequently used as input to infer language trees via standard phylogenetic methods. Lexical data are typically cognate data that rely upon concept or meaning lists, such as the Swadesh List \cite{swadesh55}, for instance. When assembling data for a specific language, linguists typically attempt to identify a frequently used every-day word for each concept, which describes it as precisely as possible \cite{dunn13}. This data assembly process induces intrinsic difficulties, due to the inherent subjectivity in concept interpretation \cite{list18}. Further, there often exist synonyms, that is, multiple words describe the same concept in a language and are used interchangeably by speakers. In German, for example, there are two words describing the concept "to kill": "töten" (related to the English word "dead") and "umbringen" (related to the English word "to bring"). The intricate nuances in meaning and usage are hard to determine and quantify \cite{list18}. In the early days of language phylogenetics, Swadesh recommended to use the most common word only, "avoiding the complication of having to deal with a choice" \cite{swadesh51}. In "The ABC's of Lexicostatistics" \cite{gudschinsky56}, Gudschinsky advises to ensure objectivity by tossing a coin to decide which word to pick when there are several choices. 

Computational phylogenetic methods have recently been applied to several cognate datasets. The inferences have mainly been conducted using Bayesian Inference (BI) methods \cite{kolipakam18, sagart19, heggarty23}, but a publication by \cite{jaeger18} shows that Maximum Likelihood (ML) based tree inference is useful as well, especially on extremely large language trees. For ML and BI approaches, that both heavily rely on the same type of phylogenetic likelihood calculations, cognate data are typically encoded via binary character matrices that represent the complete dataset including {\em all} synonyms. A question that has not been addressed to date is whether phylogenetic likelihood models as used in standard ML and BI tree inference can accommodate this data representation or whether it is preferable to choose synonyms via a labor-intensive, potentially error-prone, and subjective manual process beforehand.

Here, we focus on Maximum Likelihood (ML) tree inferences using the widely used RAxML-NG tool. Initially, working with empirical language data, we show that the topology of a tree inferred for a cognate dataset containing selected synonyms only can differ by up to 100\% from the tree topology inferred for the corresponding complete dataset including all synonyms. Given these large potential discrepancies, we advise against manual selection.
To alleviate this issue, we explore the potential benefits of using two types of alternative character matrices beyond the standard binary ones. The character matrices we propose can seamlessly be read as input by RAxML-NG while representing the complete dataset including all synonyms. We analyze the tree topologies resulting from ML inferences on all three character matrix types. We find that it depends on the respective dataset for which character matrix type the inferred tree best corresponds to the gold standard.

The remainder of this paper is structured as follows: First, we introduce our materials and methods. In particular, we formally define cognate data and describe the assembly process of the different character matrix types. Then, we evaluate how synonym selection affects the ML tree topologies inferred with RAxML-NG. Finally, we compare the introduced character matrix types. We consider the trees inferred with RAxML-NG and assess the matrix types based on how close the respective inferred trees are to the Glottolog gold standard.

\section{Materials and Methods}
\subsection{Cognate Data}
\label{sec:cognate-data}

Each cognate dataset is based on a list of concepts. Collecting data for the languages under study results in an assignment of a set of words to each language-concept pair. 
From these data, we construct a matrix $M$ containing the words' cognate classes. Cognate classes unite words that have been derived from a common ancestor \cite{dunn13} (see \cref{fig:msa-example}). We assume that the concept lists are reasonably assembled, that is, there exists at least one word for each concept in each language. When no word is given for a language-concept pair in the original data we interpret this as missing information. \\
If there are multiple synonym words describing a concept in a language, we say that the respective matrix cell is a \textit{multi-state cell}. Otherwise, if there only is a single word describing a concept in a language, we call the respective matrix cell a \textit{single-state cell}.

\begin{figure}[h]
\input{msa_example.tex}
\caption{(a): Native cognate data (b): Corresponding matrix $M$ with cognate classes (b)}
\label{fig:msa-example}
\end{figure}

\subsection{Character Matrix Types}
\subsubsection{Binary Character Matrices}

A cognate dataset can be represented by a binary character matrix $A^{\bin}$ containing the symbols $\mathtt{0}$ and $\mathtt{1})$. Additionally, specific entries may be set to the undetermined character $\mathtt{-}$, to represent missing information. Given $\mathtt{-}$ at a certain column for a language, this language does not contribute anything to the respective per-column likelihood score in RAxML-NG. Hence, the missing entries do not affect the inference. However, the lack of information itself may impact the results \cite{roure12}.
\newpage
We obtain $A^{\bin}$ as the presence-absence-matrix corresponding to the matrix containing the cognate classes (see \cref{fig:msa-exampled-deterministic} (a)). Each concept is therefore represented by as many columns as there are cognate classes, each corresponding to a specific cognate class. If there exists a word belonging to this cognate class in a certain language, the respective entry is set to $\mathtt{1}$, and to $\mathtt{0}$ otherwise. Thereby we assume that for each concept, there exists at least one word in every language. If there is no cognate class provided for a languages and a concept, this corresponds to missing information. We consequently set all columns corresponding to this concept to $\mathtt{-}$.

\subsubsection{Multi-valued Character Matrices}
Some cognate datasets can also be represented by a multi-valued character matrix $A^{\multi}$ containing multiple distinct symbols. In RAxML-NG, multi-valued character matrices are restricted to a maximum of 64 distinct symbols \cite{kozlov19}. In $A^{\multi}$, each concept is represented via a single data column only, but a different symbol is used for each cognate class. Multi-valued character matrices are thus restricted to represent a single cognate class for each language-concept pair. In order to construct $A^{\multi}$, the matrix $M$ describing the respective cognate dataset must therefore contain single-state cells only. As this is generally not the case, we exclude multi-valued character matrices from further considerations.

\begin{figure}[h]
\input{msa_example_deterministic.tex}
\caption{(a): Binary character matrix $A^{\bin}$ (b): Multi-valued character matrix $A^{\multi}$, both corresponding to the cognate dataset in \cref{fig:msa-example}. Note that $A^{\multi}$ is invalid, because $M($English, big$)$ is a multi-state cell.}
\label{fig:msa-exampled-deterministic}
\end{figure}

\subsubsection{Probabilistic Character Matrices}
\begin{figure}[h]
\centering
\includegraphics[width=0.49\textwidth]{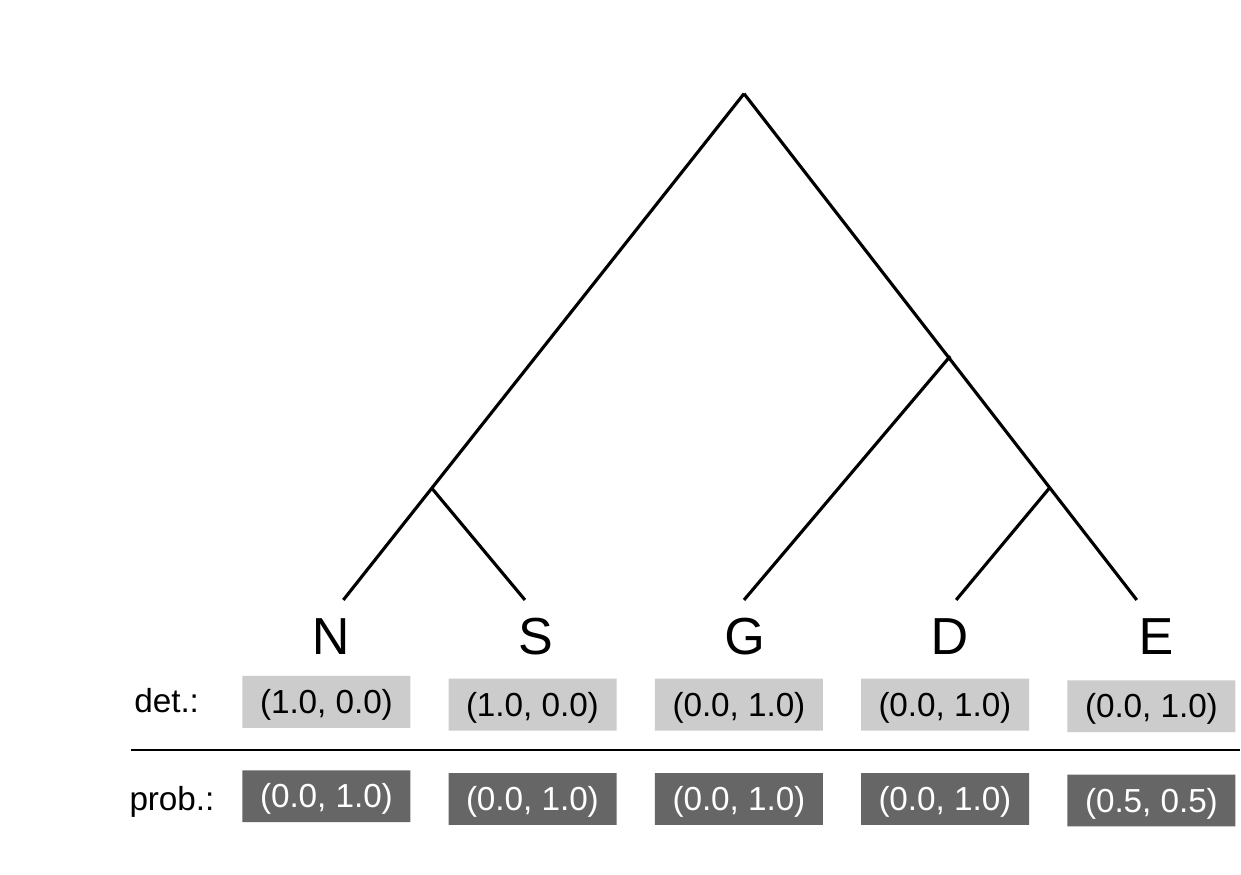}
\caption{Tree with conditional likelihood tip vectors for the per-column likelihood of the column big\_2. The light gray vectors refer to the inference on $A^{\bin}$ (\cref{fig:msa-exampled-deterministic}), the dark gray ones to the inference on  $A^{\pbin}$ (\cref{fig:msa-example-probabilistic}).}
\label{fig:prob-model}
\end{figure}

The character matrices described so far are all deterministic, because we assume that a fixed symbol is observed at each data column for each language. In a probabilistic character matrix, we instead assume that distinct symbols can occur with a certain probability, which is provided in the matrix. To represent missing data we explicitly set the probabilities for all symbols to $1.0$ \cite{kozlov19}. This encoding does not provide any information, and hence, the missing entries do not contribute to the likelihood score.\\
We can represent a probabilistic character matrix in a file via the so-called CATG-Format that is supported by RAxML-NG \cite{kozlov19}. A tree inference based on a probabilistic character matrix differs from a standard inference with respect to the form of the conditional likelihood vectors at the tips. Usually, such a vector contains a single 1.0 entry for the observed discrete character while the remaining entries are all set to 0.0. In contrast, for a probabilistic character matrix, the conditional likelihood vectors are determined based on the provided probabilities \cite{kozlov19} (see \cref{fig:prob-model}).

When interpreting cognate data in a probabilistic manner, we can represent the datasets via probabilistic character matrices. If $k$ synonyms exist for a concept in a language, we can assume that each of them occurs with probability $\frac{1}{k}$. Based on these probabilities, we can subsequently assemble probabilistic binary and multi-valued character matrices. 
In the probabilistic binary character matrix $A^{\pbin}$, a concept is represented by as many columns as there are cognate classes, just as in the corresponding deterministic binary character matrix $A^{\bin}$. At a column corresponding to one of the $k$ synonyms, we observe the symbol $\mathtt{1}$ with probability $\frac{1}{k}$ and the symbol $\mathtt{0}$ with probability $1 - \frac{1}{k}$ for the respective language.
In the multi-valued probabilistic character matrix $A^{\pmulti}$, each cognate class related to a certain concept is encoded by a different symbol but the concept is represented by only one single column. At this column, we observe each of the symbols representing one of the $k$ synonyms with probability $\frac{1}{k}$ (see \cref{fig:msa-example-probabilistic}).

\begin{figure}[h]
\input{msa_example_probabilistic.tex}
\caption{(a): Probabilistic binary character matrix $A^{\bin}$ (b): Probabilistic multi-valued character matrix $A^{\multi}$, both corresponding to the cognate dataset in \cref{fig:msa-example}.}
\label{fig:msa-example-probabilistic}
\end{figure}

\subsection{Comparing Trees}
We measure topological dissimilarities between inferred phylogenetic trees using the \textit{Robinson-Foulds (RF) distance} \cite{robinson81}. This standard metric is based on non-trivial splits in trees. A split is a partitioning of the tree's tips into two sets corresponding to the subtrees that arise when a branch of the tree is removed. A split is called non-trivial, if each set contains at least two tips. The absolute RF distance between two trees is defined as the number of non-trivial splits, which are unique to one of the two trees. In the following, we use the relative RF distance which we obtain by normalizing the absolute RF distance with $2(n - 3)$, the total number of non-trivial splits in two strictly binary unrooted trees.\\
The inferred ML trees are strictly binary, but polytomies can occur in manually constructed reference trees (e.g., in Glottolog reference trees, see below). To compare an inferred ML tree to a reference tree, we therefore use the \textit{generalized quartet (GQ) distance} \cite{pompei11} instead. This metric has the advantage that it yields a distance of $0$ if there are no contradictions between the inferred tree and the reference tree, even if the reference tree does contain polytomies. To calculate the GQ distance, we extract all possible quartets of tips in the tree. For each quartet, we then determine the topology of the 4-tip subtree induced by the comprehensive tree. When comparing two trees, the GQ reflects the proportion of quartets that induce distinct topologies. Note that the RF distance and the GQ distance are distributed differently \cite{steel93}, which is illustrated by the following example: Let $T$ be a fully balanced strictly binary tree with $16$ leaves and let further $T^{\prime}$ be a tree obtained from $T$ by swapping two leaves from neighboring subtrees (for details refer to \cref{fig:mini-example} in the supplement). The RF distance of $T$ and $T^{\prime}$ is $0.15$, while they exhibit a GQ distance of $0.03$ only.

\subsection{Maximum Likelihood Tree Inferences}
In our experiments, we execute 20 independent ML tree searches on each character matrix under study. We use the default tree search configuration of RAxML-NG (10 searches starting from random trees and 10 searches starting from randomized stepwise addition order parsimony trees). For the inferences on both, the deterministic, and the probabilistic binary character matrices, we use the BIN+G model of binary character substitution. For the tree searches on probabilistic multi-valued character matrices we use the MK+G model which allows using up to 64 different characters but assumes equal substitution rates between all characters. Using BIN+G (MK+G resp.), we approximate the $\Gamma$ model of rate heterogeneity via four discrete rates. Thus, each inference includes the ML estimation of the $\alpha \in [0, 100]$ shape parameter that determines the shape of the $\Gamma$ distribution. The smaller the estimate of $\alpha$, the higher the rate heterogeneity in the respective dataset will be \cite{yang95}. 
The command lines we used to execute the inferences are available on Github (\url{https://github.com/luisevonderwiese/synonyms}). 

\subsection{Quantifying Difficulty}
To quantify the difficulty of a phylogenetic inference for a given dataset we use the difficulty score as predicted by Pythia \cite{haag22}. The tool internally uses a Random Forest Regressor \cite{ho95} to predict this difficulty score based on attributes of the character matrices and on the results of computationally substantially less expensive parsimony-based tree inferences \cite{farris70}. Because the parsimony approach can only be applied to deterministic character matrices, the difficulty score is also limited to this matrix type.

\subsection{Data}
For our analyses we use 44 cognate datasets. We retrieve the vast majority (39 datasets) from the cross-linguistic lexical database \textit{Lexibank} \cite{list22}. The five remaining datasets originate from the supplementary material provided for the book "Sequence Comparison in Historical Linguistics" \cite{list21}. 
The above repositories comprise more datasets than we use here, as not all of them are suitable for our experiments. Since GQ distances can only be calculated on trees with strictly more than 4 tips, we exclude datasets comprising less than 5 languages. 
We also do not consider datasets with more than 400 languages due to the excessive tree inference times. In addition, these datasets exhibit an unfavorable number of concepts to number of languages ratio which yield them difficult to reliably infer. We therefore do not expect the respective tree inference results to be informative. We further exclude datasets with regard to the maximum number of different cognate classes that are occurring for the concepts. If only one cognate class is used for each concept, we do not consider the respective dataset, as the corresponding character matrices are not informative. We also exclude datasets comprising concepts with more than 64 distinct cognate classes, as no probabilistic multi-valued character matrix can be constructed in this case, because RAxML-NG is limited to using a maximum of 64 distinct symbols.\\
As gold standard, we use the manually constructed tree published by \citex{hammarstroem22} in the \textit{Glottolog} database. This tree contains all $8205$ languages listed in Glottolog. For each dataset, we obtain the respective gold standard tree by constraining the comprehensive tree to the languages contained in the dataset by pruning the languages not contained therein. For a dataset to be suitable for our experiments, it must be possible to extract an informative reference tree. To this end, we exclude datasets, if the corresponding reference tree has a star topology because comparisons of binary topologies with a star topology do not yield meaningful topological distances.\\
In both data sources, the datasets are standardized as specified by the Cross-Linguistic Data Format (CLDF) \cite{forkel18}. Our implementation for converting CLDF data into the character matrix types we describe here is available on Github (\url{https://github.com/luisevonderwiese/lingdata}).

\section{Results}
\subsection{Effects of Synonym Selection}
\begin{figure}[h!]
\centering
\includegraphics[width=0.49\textwidth]{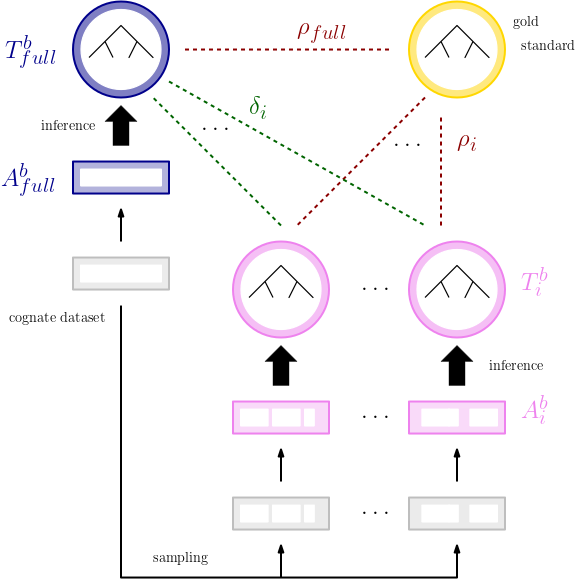}
\caption{Experimental setup for assessing the effects of synonym selection: For each dataset, we create $100$ selection samples and construct the corresponding (deterministic) binary character matrices $A^{\bin}_i$, $i \in {1, \dots, 100}$. $A^{\bin}_{\full}$ represents the complete dataset including all synonyms. For each character matrix we consider the best scoring tree resulting from 20 independent tree searches with RAxML-NG, denoted by $T_i$, $i \in {1, \dots, 100}$, $T_{\full}$ respectively. $\delta_i$ corresponds to the RF distance between $T_i$ and $T_{\full}$, $\rho_i$ to the GQ distance between $T_i$ and the gold standard tree. By $\rho_{\full}$ we denote the GQ distance between $T_{\full}$ and the gold standard.}
\label{fig:experiment1}
\end{figure}

In the following, we aim to assess, whether ML tree inferences are feasible on binary character matrices representing cognate datasets with {\em all} synonyms or whether it is preferable to choose synonyms manually in advance. For this purpose, we investigate how it impacts the results of RAxML-NG based tree inferences when different combinations of synonyms are being selected. The experimental setup is illustrated in \cref{fig:experiment1}\\
\newpage
\begin{figure}[h!]
\centering
\includegraphics[width=0.49\textwidth]{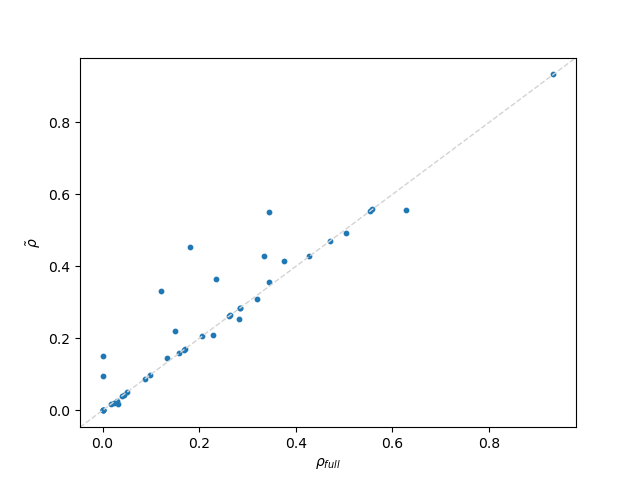}
\caption{Each marker corresponds to a specific dataset under study. The x-axis indicates $\rho_{\full}$, the GQ distance of the respective tree $T_{\full}$ to the gold standard. The y-axis indicates $\tilde{\rho}$, the median of the GQ distances of the trees $T_i$ to the gold standard. 
For the majority of the datasets, the tree $T_{\full}$ is at least as close to the gold standard as the median of the trees $T_i$.
}
\label{fig:scatter-median}
\end{figure}
We generate selection samples from each examined dataset by selecting synonyms uniformly at random. Hence, if $k$ synonyms exist for a concept in a language, we pick one of them with probability $\frac{1}{k}$. We create 100 such selection samples for each dataset under study. For some of the datasets, we performed the following experiments with 1000 samples, but this did not lead to substantial differences in the results. For each sample we construct a (deterministic) binary character matrix $A^{\bin}_i$, $i \in {1, \dots, 100}$. Note that each of these matrices only contains information about the selected synonyms. Then, using RAxML-NG with the tree search options as outlined above we execute 20 independent ML tree searches on each of these character matrices as well as on the character matrix $A^{\bin}_{\full}$ representing the complete dataset including all synonyms. For each character matrix $A^{\bin}_i$, we consider the best-scoring ML tree $T_i$ we inferred on it. Let further $T_{\full}$ be the best-scoring tree resulting for $A^{\bin}_{\full}$.
For each dataset, we use the corresponding gold standard tree from Glottolog as a reference. Let $\rho_{\full}$ be its GQ distance to $T_{\full}$. For each tree $T_i$, we denote its GQ distance to the gold standard by $\rho_i$. Let further $\tilde{\rho}$ be the median of the GQ distances $\rho_i$. For $\rho_{\full}$, we obtain an average distance of $0.22$ over all 44 datasets while $\tilde{\rho}$ averages to $0.25$. Thus, the two approaches appear to perform equally well at first sight, with $T_{\full}$ being only slightly closer to the reference, on average. \\

For a more detailed assessment, we compare $\rho_{\full}$ and $\tilde{\rho}$ for each dataset (see \cref{fig:scatter-median}). For 33 out of 44 examined datasets, we observe that $\rho_{\full} \leq \tilde{\rho}$, that is $T_{\full}$ comes closer to the gold standard. In most of the cases, the inference on $A^{\bin}_{\full}$ thus performs better than the median inference on the sampled character matrices. For the datasets where $\rho_{\full} > \tilde{\rho}$ applies, the difference never exceeds $0.07$ and only for 5 datasets it exceeds $0.01$. If the median tree $T_i$ is closer to the gold standard, the differences are hence not substantial in most cases. These results speak in favor of performing inferences on the full dataset as the results tend to be slightly better than for the median randomized synonym selection. When using the mean GQ distance instead of the median, the observations are analogous. For details refer to \cref{fig:scatter-mean} in the supplement. \\
\begin{figure}[h!]
\centering
\includegraphics[width=0.49\textwidth]{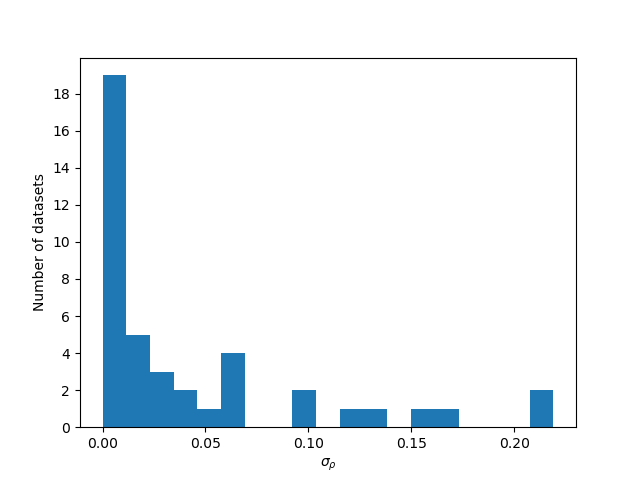}
\caption{Distribution of $\sigma_{\rho}$, the standard deviation of the GQ distances of the trees $T_i$ to the gold standard.\\
The x-axis indicates $\sigma_{\rho}$, the y-axis the number of datasets such that respective standard deviation occurs. The observed standard deviations indicate that the inferred trees are substantially different depending on the subset of selected synonyms.}
\label{fig:hist-gqd-sampled-std}
\end{figure}

The advantages of using the full datasets are further emphasized when analyzing the stability of the tree inference under the described sampling procedure.
For this purpose, we evaluate how much the trees $T_i$ vary in terms of their deviation from the gold standard. For a fixed dataset, let $\sigma_{\rho}$ be the standard deviation of the GQ distances $\rho_i$. The distribution of $\sigma_{\rho}$ over all examined datasets is depicted in \cref{fig:hist-gqd-sampled-std}. For $13$ datasets, we observe $\sigma_{\rho} > 0.05$, for $2$ of them, even $\sigma_{\rho} > 0.2$. Since the GQ distance is distributed differently than the RF distance, the observed standard deviations indicate substantial differences. How close the inferred tree comes to the gold standard can therefore vary considerably depending on the subset of selected synonyms.\\

\begin{figure}[h!]
\centering
\includegraphics[width=0.49\textwidth]{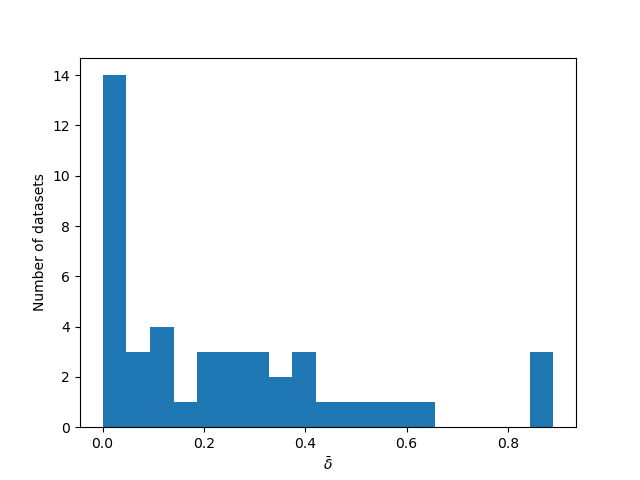}
\caption{Distribution of $\bar{\delta} \coloneqq (\sum_{i=1}^{100} \delta_i) / 100$ \\
The x-axis indicates $\bar{\delta}$, the y-axis the number of datasets such that the trees $T_i$ yield the respective average RF distance to $T_{\full}$. The lower $\bar{\delta}$, the more stable the respective dataset is under random synonym selection.}
\label{fig:hist-rf-bin-avg}
\end{figure}

In an additional stability analysis, we examine, to which extent the trees $T_i$ deviate from the respective tree $T_{\full}$. For each tree $T_i$, we therefore determine $\delta_i$ as its RF distance to $T_{\full}$. We consider $\bar{\delta} \coloneqq (\sum_{i=1}^{100} \delta_i) / 100$, indicated on the x-axis of \cref{fig:hist-rf-bin-avg}. The figure's y-axis gives the number of examined datasets for which the respective average distance results from the described experiment. $\bar{\delta}$ is close to $0$ for some datasets but also $\geq 0.4$ or even $\geq 0.8$ for a considerable proportion of datasets. $\bar{\delta}$ can be interpreted as an indication for the stability of the inferred tree under sampling. The lower $\bar{\delta}$, the more stable the respective dataset is.\\

For each dataset we also determine the difficulty of the inference on the character matrix $A^{\bin}_{\full}$. The obtained difficulty score is only slightly correlated with the stability as quantified by $\bar{\delta}$ (Pearson correlation $0.43$, P-value $0.003$). 
Further, we examine the proportion of multi-state cells for each cognate dataset, corresponding to the proportion of language-concept pairs such that there are words from more than one cognate class describing the concept in the language. This score is slightly correlated with $\bar{\delta}$ (Pearson correlation $0.45$, P-value $0.002$). The more multi-state cells exist, the more information are discarded during sampling and the more the sampled character matrix differs from the character matrix that is based on the full dataset, leading to the observed correlation.\\

\begin{figure}[h!]
\centering
\includegraphics[width=0.49\textwidth]{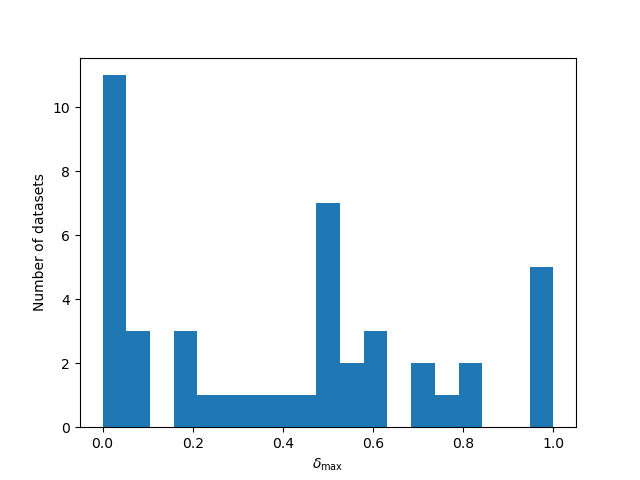}
\caption{Distribution of $\delta_{\max} \coloneqq \max({\delta_i : i \in {1, \dots, 100}})$.\\
The x-axis indicates $\bar{\delta}$, the y-axis the number of datasets such that respective maximum value occurs among the RF distances of the trees $T_i$ to $T_{\full}$. The analysis illustrates that there exist datasets for which the worst case synonym choice leads to a tree which is entirely different (RF distance of 1.0) from the tree inferred on the dataset including all synonyms.}
\label{fig:hist-rf-bin-max}
\end{figure}
We finally consider $\delta_{\max} \coloneqq \max({\delta_i : i \in {1, \dots, 100}})$. The analysis of this score elucidates the detrimental effect of an unfavorable worst case synonym choice. \cref{fig:hist-rf-bin-max} depicts the respective distribution for the datasets under study. We observe that there is a considerable proportion of datasets with $\delta_{\max} \geq 0.5$. For $5$ datasets, we even observe $\delta_{\max} = 1$. For these datasets there exists at least one character matrix containing a certain subset of synonyms such that the inferred tree admits an RF distance of $1.0$ to the tree resulting from the full dataset.\\

Our observations illustrate that the synonym selection can induce entirely different RAxML-NG based ML tree topologies. The decision, which synonyms to consider thus substantially affects the results. Therefore, we strongly advise against manual synonym selection. Instead, we recommend to consider all known synonyms when inferring phylogenetic trees. Our analysis shows, that tree inference on the respective representation as a binary character matrix leads to feasible results. Additionally, this circumvents the labor-intensive process of selecting synonyms manually. 
\subsection{Modeling Data with Synonyms}
\begin{figure}[h!]
\centering
\includegraphics[width=0.49\textwidth]{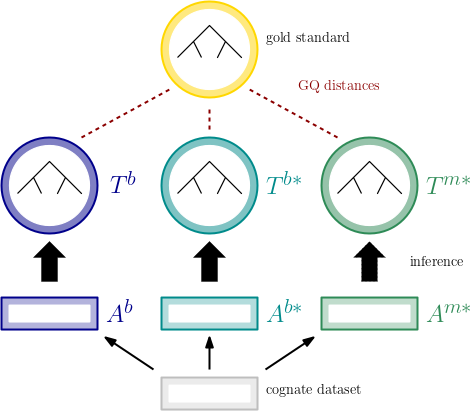}
\caption{Experimental setup for comparing the performance of different character matrix types: For each dataset we construct a deterministic binary character matrix $A^{\bin}$, a probabilistic binary character matrix $A^{\pbin}$, and a probabilistic multi-valued character matrix $A^{\pmulti}$. For each character matrix type, we consider the best tree scoring tree resulting from 20 independent tree searches with RAxML-NG, denoted by $T^{\bin}$, $T^{\pbin}$, $T^{\pmulti}$ respectively. We compute the GQ distances of these trees to the gold standard from Glottolog.}
\label{fig:experiment2}
\end{figure}

In the following, we compare the performance of ML tree inferences on three different kinds of character matrices representing cognate data. For each dataset under study, we consider its representation as a deterministic binary character matrix $A^{\bin}$, as a probabilistic binary character matrix $A^{\pbin}$, and as a probabilistic multi-valued character matrix $A^{\pmulti}$. On each character matrix type we execute 20 independent ML tree searches with RAxML-NG as described above. The structure of the following experiment is illustrated in \cref{fig:experiment2}. \\
We aim to assess how suitable the different character matrix types are for capturing the signal contained in the data during ML tree inference with RAxML-NG. To this end, we examine the resulting trees. For a fixed dataset, let $T^{\bin}$, $T^{\pbin}$, $T^{\pmulti}$ be the best-scoring tree inferred on the respective character matrix type. We compare these trees to the corresponding gold standard from Glottolog. We are interested in which character matrix type will induce the tree that is closest to it. We henceforth say that this is the character matrix type performing best.\\

The trees inferred on $A^{\bin}$ yield an average GQ distance of $0.22$ to the respective gold standard tree. For both probabilistic character matrix types we observe an average GQ distance of $0.23$. At first glance, all character matrix types therefore appear to yield results of comparable quality. For $9$ datasets, $A^{\bin}$ performs best, for $9$ datasets it is $A^{\pbin}$, and for $11$ datasets $A^{\pmulti}$. In $8$ datasets, all character matrix types perform equally well. There are $7$ datasets where two distinct character matrix types yield equally good results, but are better than the third. For the sake of simplicity, we exclude these datasets from our further analyses. There is no clear trend for one character matrix type always being preferable over the others. Conversely, based on our results, we cannot advise against the use of any type of character matrix.\\

Subsequently, we analyze the differences between the trees inferred on the different character matrix types, aiming to show that these differences are so substantial that it is indeed worthwhile to perform inferences on all three character matrix types. 
For this purpose we first consider the datasets, for which $A^{\bin}$ performs best. Over these datasets, $T^{\bin}$, which comes closest to the gold standard tree, yields an average RF distance of $0.60$ to $T_d^{\pbin}$ and of $0.61$ to $T_d^{\pmulti}$. This indicates that the trees resulting from the inferences on the various character matrix types can differ considerably. We observe an analogous behavior for the datasets where $A^{\pbin}$ or $A^{\pmulti}$ perform best (see \cref{tab:comp-rf} in the supplement). We conclude that the differences between the inferred trees can become so large such that all three character matrix types always need to be considered.\\

In the following, we attempt to identify properties, that for a given dataset, might be able to predict, which character matrix type performs best.
First, we consider the $\alpha$ shape parameter that determines the shape of the $\Gamma$ distribution under the BIN+G (MK+G resp.) model. For $9$ out of the $11$ datasets with $A^{\pmulti}$ performing best, the ML estimates of $\alpha$ are below $20$, indicating a moderate to high degree of rate heterogeneity. This is also the case for $8$ out of the $9$ datasets for which $A^{\pbin}$ performs best. However, $\alpha < 20$ is only observed for $5$ out of the $9$ datasets for which the binary character matrix yields the best performance and only for $4$ out of the $8$ datasets for which all character matrix types perform equally well.
We therefore observe a tendency for probabilistic modeling to be advantageous for datasets with high rate heterogeneity. These datasets may exhibit a larger variance with respect to the number of cognate classes per concept, which can possibly be better accommodated via a probabilistic character matrix type.\\

We further investigate whether the difficulty of the ML inference on a certain dataset is related to character matrix performance. In the following, the difficulty reported for a dataset is the difficulty obtained for $A^{\bin}$, as difficulty scores can only be computed for deterministic character matrices. The datasets for which all character matrix types perform equally well exhibit a comparably low average difficulty of $0.17$. A low difficulty score indicates a strong phylogenetic signal in the data. This strong signal can be captured during the ML tree inference, regardless of the type of character matrix used to represent it. The datasets with $A^{\bin}$ performing best exhibit an average difficulty of $0.45$. For the datasets for which $A^{\pbin}$ performs best, the average difficulty is $0.29$, and for those for which $A^{\pmulti}$ yields the best performance, it amounts to $0.18$. While the probabilistic character matrix types are hence advantageous for data with a clear signal, the deterministic representation is better suited to capture the signal in datasets where this is more difficult.

\section{Conclusion}

We demonstrated, that the selection of synonyms in cognate datasets {\em can} induce substantially different tree topologies when performing ML inferences with RAxML-NG. It is thus preferable to perform tree inferences on the full dataset with all synonyms included. This also circumvents the labor-intensive process of manual synonym selection. The datasets can be encoded as (deterministic) binary character matrices. In addition, we introduced probabilistic binary and probabilistic multi-valued character matrices as alternative representations. We showed that it is dataset-dependent, for which character matrix type the inferred tree is closest to the gold standard. We were able to identify the rate heterogeneity and the difficulty score as properties that may indicate which character matrix type is best suited for a given dataset. Note that unfortunately, the number of available cognate datasets is too low in order to train any machine-learning based predictors. We therefore recommend performing inferences on all three character matrix types when analyzing cognate datasets. We provide a Python interface on Github (\url{https://github.com/luisevonderwiese/lingdata}) that can be used to create all of the above character matrices for any cognate dataset provided in CLDF format. 

\section{Future Work}
Our work leads to several novel questions that need to be addressed. When constructing probabilistic character matrices, data from corpus studies could be taken into account instead of assuming a uniform probability distribution of synonym occurrence, albeit this information could be challenging to obtain for small languages and dialects. Future work should further strive to develop an alternative model of evolution taking idiosyncrasies of language data into account. Another open question is how the quality of the inference methods can be assessed without referring to the gold standard. This requires the development of language-specific data simulation tools, taking into account the challenges that have been described with respect to simulating realistic DNA data \cite{trost23}.\\

\section*{Acknowledgement}
Luise Häuser and Alexandros Stamatakis are financially supported by the Klaus Tschira Foundation, and by the European Union (EU) under Grant Agreement No 101087081 (CompBiodiv-GR). Gerhard Jäger is supported by the ERC-AdG 834050 (CrossLingference) and the DFG-FOR 2234 \textit{Words, Bones, Genes, Tools}.
\\

\includegraphics[width=0.49\textwidth]{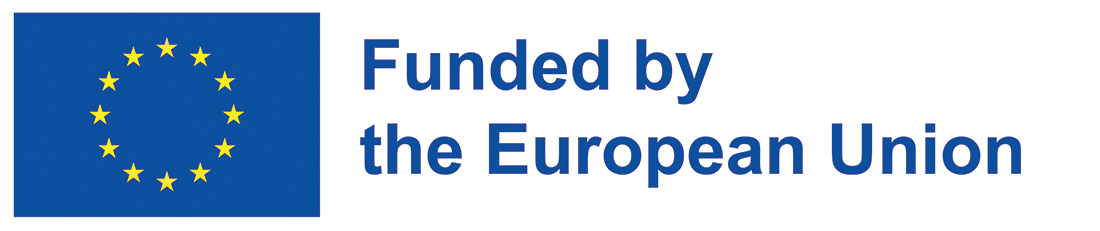}

\newpage
\bibliography{references}

\newpage
\FloatBarrier

\appendix

\section{Supplementary Figures and Tables}
\label{sec:appendix}
\begin{figure}[h!]
\centering
\begin{subfigure}[b]{0.49\textwidth}
   \includegraphics[width=1\linewidth]{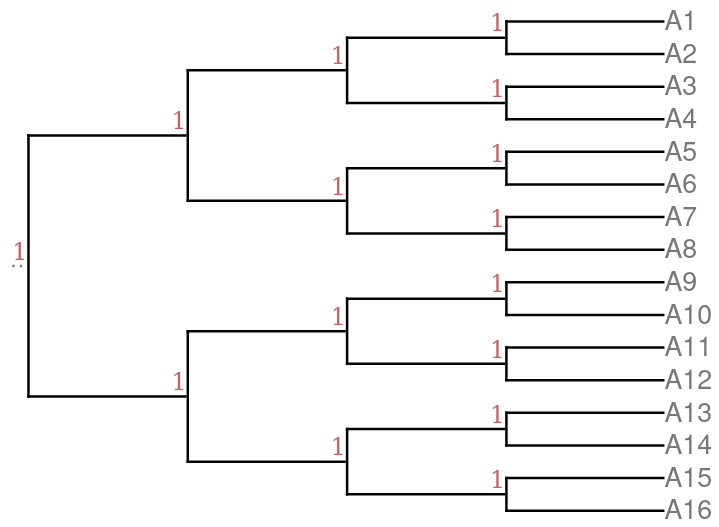}
   \caption{Tree $T$, fully balanced strictly binary tree with $16$ leaves}
\end{subfigure}

\begin{subfigure}[b]{0.49\textwidth}
   \includegraphics[width=1\linewidth]{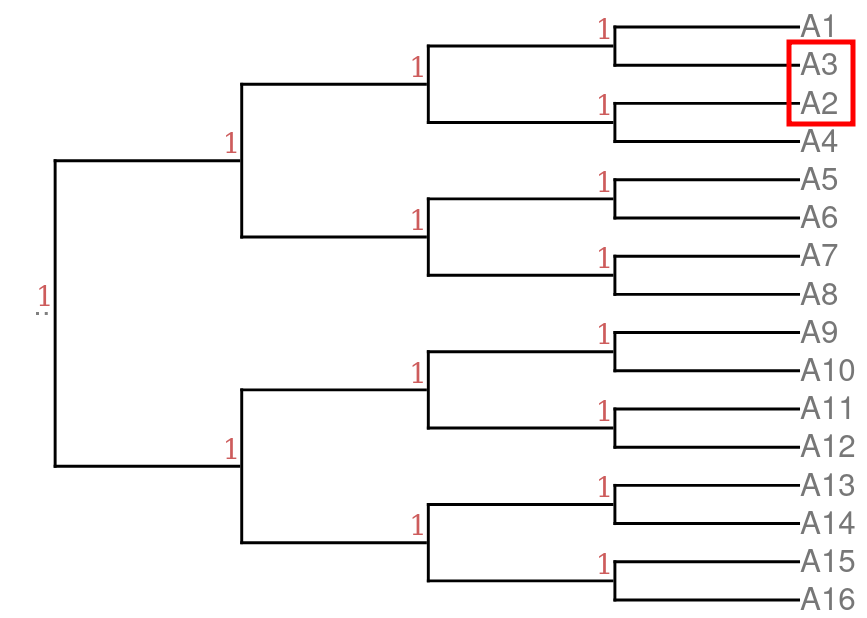}
   \caption{Tree $T^{\prime}$, differences to $T$ are highlighted in red}
\end{subfigure}
\caption{The trees $T$ and $T^{\prime}$ exhibit an RF distance of $0.15$ but a GQ distance of $0.03$, which illustrates the different distributions of the metrics.}
\label{fig:mini-example} 
\end{figure}
\phantom{
d\\
d\\
d\\
d\\
d\\
d\\
d\\
d\\
d\\
d\\
d\\
d\\
d\\
d\\
d\\
d\\
}
\newpage
\begin{figure}[h!]
\centering
\includegraphics[width=0.49\textwidth]{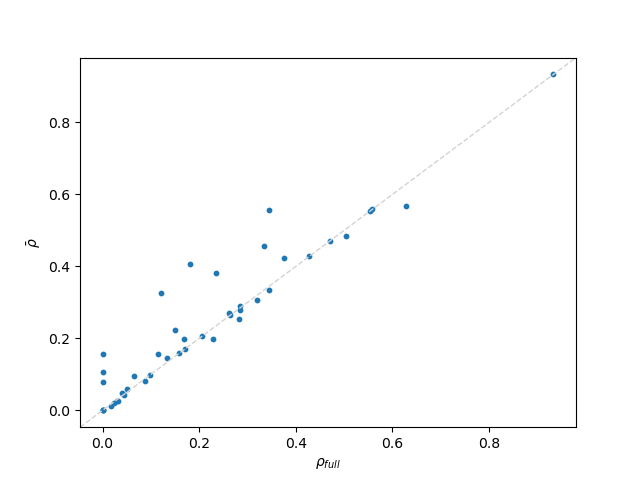}
\caption{Each marker corresponds to a specific dataset under study. The x-axis indicates $\rho_{\full}$, the GQ distance of the respective tree $T_{\full}$ to the gold standard. The y-axis indicates $\bar{\rho}$, the average GQ distance of the trees $T_i$ to the gold standard.Most markers are located on the identity (represented by a dashed line) or above, that is the tree $T_{\full}$ is at least as close to the gold standard as the average tree $T_i$ for the majority of the datasets.}
\label{fig:scatter-mean}
\end{figure}

\renewcommand{\arraystretch}{1.2}
\FloatBarrier
\begin{table}[h!]

\begin{tabular}{|l|l|l|l|} 
\hline
Mean $\rf(\cdot, \cdot)$ & $T^{\bin}$ & $T^{\pbin}$ & $T^{\pmulti}$ \\
\hline
\multicolumn{4}{|l|}{Datasets with $A^{\bin}$ performing best:} \\
\hline
$T^{\bin}$ & $\mathbin{\color{gray}0.00}$ & $0.60$ & $0.61$ \\
\hline
\multicolumn{4}{|l|}{Datasets with $A^{\pbin}$ performing best:} \\
\hline
$T^{\pbin}$ & $0.25$ & $\mathbin{\color{gray}0.00}$& $0.36$ \\
\hline
\multicolumn{4}{|l|}{Datasets with $A^{\pmulti}$ performing best:} \\
\hline
$T^{\pmulti}$ & $0.43$ & $0.40$ & $\mathbin{\color{gray}0.00}$ \\
\hline
\end{tabular}

\caption{Mean RF distances among the trees resulting from the inferences on the different character matrix types. Datasets are grouped according to the character matrix type performing best. The differences between the trees are so substantial that it is indeed worthwhile to perform inferences on all three character matrix types.}
\label{tab:comp-rf}
\end{table}

\end{document}